%% file: main.tex
\pdfoutput=1

\documentclass[11pt]{article}

\usepackage[]{EMNLP2023}

\usepackage{times}
\usepackage{latexsym}

\usepackage[T1]{fontenc}

\usepackage[utf8]{inputenc}

\usepackage{microtype}

\usepackage{inconsolata}

\usepackage{booktabs}
\usepackage{tkz-kiviat}
\usepackage{xspace}
\usepackage{amsmath}

\usepackage{graphicx}
\usepackage{caption}
\usepackage{subcaption}
\usepackage{float}
\usepackage{stfloats}
\usepackage{tikz}
\usepackage{nicefrac}
\usepackage{pgfplots}
\usepgfplotslibrary{fillbetween}
\usetikzlibrary{patterns}
\usetikzlibrary{shapes.geometric}
\pgfplotsset{compat=1.6}

\def\addlegendimage{\csname pgfplots@addlegendimage\endcsname}

\input{custom_commands}

\title{\memoryvq: Compression for Tractable Internet-Scale Memory}

\author{
Yury Zemlyanskiy\thanks{\- \ Equal contribution. Correspondence to \{yury,michiel\}@augmentcode.com} \ \footnotemark[2]\thanks{\- \ Augment Computing. Work done at Google Research.} ,~ Michiel de Jong\footnotemark[1] \ \footnotemark[2] ,~ {\bf Luke Vilnis} \\ {\bf Santiago Onta\~{n}\'{o}n},~ {\bf William W. Cohen},~ {\bf Sumit Sanghai},~ {\bf Joshua Ainslie}
  \AND
  {\rm \Large Google Research}\\
  }

\begin{document}
\maketitle
\begin{abstract}
Retrieval augmentation is a powerful but expensive method to make language models more knowledgeable about the world. Memory-based methods like \lumen \cite{lumen} pre-compute token representations for retrieved passages to drastically speed up inference. However, memory also leads to much greater storage requirements from storing pre-computed representations.

We propose \modelname, a new method to reduce storage requirements of memory-augmented models without sacrificing performance. Our method uses a vector quantization variational autoencoder (VQ-VAE) to compress token representations. We apply \modelname to the \lumen model to obtain \lumenvq, a memory model that achieves a 16x compression rate with comparable performance on the KILT benchmark. \lumenvq enables practical retrieval augmentation even for extremely large retrieval corpora.

\end{abstract}

\section{Introduction}

Retrieval augmentation is a common method to improve the factual knowledge of language models \cite{fid, retro, rag, knnlm, realm,atlas}. Retrieval provides a model with additional context in the form of text passages relevant to an input query. However, retrieval augmentation comes at an increased computational cost, as the model must process the retrieved passages on-the-fly.

\input{results/main_table}

A recent line of work \cite{readtwice,tome,qama,fidmemory,lumen} speeds up retrieval augmentation by pre-encoding passages from the corpus in advance. This way, the model can retrieve representations instead of raw text, which avoids the cost of reading retrieved passages from scratch. One such model, \lumen, stands out for its strong performance, achieving 3x faster inference than standard Fusion-in-Decoder \cite{fid} (FiD) with minimal loss in quality.

However, these pre-encoding memory models use much more storage than traditional retrieval-augmented models - \lumen saves an embedding for each token in the corpus, which takes up much more space than token IDs. Table 1 compares storage requirements for T5 XXL-sized models. FiD requires 2 bytes to store an ID of each token, while \lumen uses a 4096-dimensional vector of \texttt{bfloat16} values, summing to 8KB per token. Wikipedia contains around 4 billion tokens, which means \lumen token representations take up 30TB. For an internet-scale corpus of 1 trillion tokens, disk requirements balloon to an impractical 7PB.

This work combines of product quantization \cite{pq} and VQ-VAE method \cite{vqvae} to significantly reduce storage requirements for memory-based methods with limited loss in quality. In particular, \lumenvq achieves a 16x compression rate, meaning we only need 2TB to store memories for all of Wikipedia and 500TB for a 1 trillion token corpus. Moreover, \lumenvq suffers minimal loss in performance on the KILT benchmark \cite{kilt} of knowledge intensive tasks.

Our contribution is the first paper on compressing pre-encoded token memory representations. This compression makes memory methods such as \lumen practical even for extremely large retrieval corpora. Previous works (e.g., \citep{colbertv2, cq, sdr, becr}) have focused on token representation compression for late-interaction reranking models. In contrast, our approach compresses the intermediate representations of a language model. These compressed representations are used as inputs into an LLM, and the compression layers' parameters are trained alongside the rest of the model.

\section{Background}
\label{section:background}

We aim to match \fid and \lumen performance in quality while reducing \lumen storage requirements. We first describe \fid and \lumen, methods on which \memoryvq is built, and their storage requirements. For an in-depth analysis, please see \citet{lumen}. We follow up with background on vector quantization, including product quantization and VQ-VAE used for \memoryvq.

\subsection{Retrieval and memory augmented models}
\subsubsection{Fusion-in-Decoder}
\label{section:fid}

Fusion-in-Decoder (FiD) \citep{fid} builds upon the T5 \citep{t5} encoder-decoder model. It retrieves relevant text passages, appends them to the input $Q$, and processes each input-passage pair with the encoder. The resulting token representations are merged and attended by the decoder. We highlight \textcolor{flycolor}{\textbf{live}} components in blue and \textcolor{precolor}{\textbf{pre-computed}} in orange. FiD does not have any pre-computed components.
\vspace{-3pt}
\begin{equation*}
    G = \text{\textbf{\textcolor{flycolor}{Dec}}}\Big[\text{\textcolor{flycolor}{\textbf{Enc}}}(Q; \text{Passage}_1); \ldots \text{\textbf{\textcolor{flycolor}{Enc}}}(Q; \text{Passage}_k)\Big]
\end{equation*}
\fid storage needs are low since we only need to store token IDs. Each ID can be encoded with 16 bits, so the storage cost for a retrieval corpus with $N$ tokens is
\vspace{-3pt}
\begin{equation*}
    S_{\text{\fid}} = 16 \cdot N
\end{equation*}
\subsubsection{\lumen}
\label{section:lumen}

\lumen \cite{lumen} reduces inference cost by partially pre-computing encoder representations for retrieved passages. Instead of retrieving the actual text, \lumen retrieves intermediate layer representations during inference.

\lumen is initialized from a pre-trained T5 encoder-decoder model, with a \textcolor{precolor}{\textbf{memory encoder}} containing the initial $1 - \liveprop$ proportion of layers and a \textcolor{flycolor}{\textbf{live encoder}} with the remaining $\liveprop$ proportion of layers. The memory encoder is applied offline to pre-compute memory representations for passages in the corpus. Later, these representations are dynamically updated with the fine-tuned live encoder based on the input and task. To ensure compatibility, \modelname applies the memory encoder to the input before concatenating the question representation with the memory representation.
\vspace{-3pt}
\begin{align*}
\label{eqn:lumen}
    & H_i = \Big[\textbf{\textcolor{flycolor}{MemEnc}}(Q);\hspace{0.2cm} \text{\textbf{\textcolor{precolor}{MemEnc}}}(\text{Passage}_i)\Big] \\
    &G = \text{\textbf{\textcolor{flycolor}{Dec}}}\Big[Q; \text{\textbf{\textcolor{flycolor}{LiveEnc}}}(H_1); \ldots \text{\textbf{\textcolor{flycolor}{LiveEnc}}}(H_k) \Big]
\end{align*} 
Choosing $\liveprop = 1$ yields a model very close to \fid while $\liveprop = 0$ is a full memory model. One of the insights of the \lumen paper is that one can match \fid performance while using small $\liveprop$, reducing inference cost to a fraction $\liveprop$ of FiD encoder FLOPs for any given model size.

\lumen keeps $d$-dimensional \textbf{\textcolor{precolor}{MemEnc}} output representations for every token. With \texttt{bfloat16} format, the total storage cost becomes
\vspace{-3pt}
\begin{equation*}
    S_{\lumen} = 16d \cdot N
\end{equation*}

\subsection{Vector quantization}
Vector quantization (VQ) is a classical compression technique for vector data. The general idea is to prepare a set of vectors known as ``codes'' and then represent each input vector with the nearest code. The approach significantly reduces storage requirements as we only need to store the integer ID of the code instead of the entire high-dimensional input vector. VQ is a lossy compression method since decompression returns the value of the nearest code (by looking up the ID) instead of the original vector. Usually, codes are generated by clustering the input vectors, for example, using \texttt{kmeans}-like methods.

\subsubsection{Product quantization}
A popular variant of vector quantization is product quantization \citep{pq,opq}. The method involves partitioning high-dimensional vectors into subspaces and independently quantizing each subspace using a vector quantization subroutine. The product quantization is frequently used in modern approximate nearest neighbor search engines \citep{scann, faiss} to speed up lookup.

\subsubsection{VQ-VAE}

The VQ-VAE approach \cite{vqvae} is a variant of variational autoencoders that utilizes vector quantization for obtaining a discrete latent representation. Notably, the VQ-VAE compression layer allows joint training with the rest of the model due to a straight-through estimator for gradient backpropagation. The method is commonly used in creating discrete representations of continuous objects such as images or audio \citep{vqvae, vqvae2}.

\section{\modelname}
We propose \memoryvq, an efficient method for reducing storage requirements for memory-based models. The high-level idea is to compress memories using vector quantization techniques and store integer codes instead of the original memory vectors. Codes are decompressed into vectors on the fly. Applying the method to \lumen yields the following \lumenvq model.
\vspace{-3pt}
\begin{align*}
\label{eqn:lumen}
    & \text{codes}_i = \text{\textbf{\textcolor{precolor}{CompressVQ}}}(\text{\textbf{\textcolor{precolor}{MemEnc}}}(\text{Passage}_i))\\
    & H_i = \Big[\textbf{\textcolor{flycolor}{MemEnc}}(Q_i);\hspace{0.2cm} \text{\textbf{\textcolor{flycolor}{DecompressVQ}}}(\text{codes}_i)\Big] \\
    &G = \text{\textbf{\textcolor{flycolor}{Dec}}}\Big[Q; \text{\textbf{\textcolor{flycolor}{LiveEnc}}}(H_1); \ldots \text{\textbf{\textcolor{flycolor}{LiveEnc}}}(H_k) \Big]
\end{align*} 
To perform \textbf{\textcolor{precolor}{CompressVQ}} and \textbf{\textcolor{flycolor}{DecompressVQ}} we apply product quantization, splitting each vector into subspaces and independently quantizing each subspace using VQ-VAE. Codes are an exponential moving average of memory vectors assigned to the code in each batch. Appendix A in \citet{vqvae} contains a detailed description.

For training the compression layer jointly with the model, we follow the VQ-VAE recipe, but we avoid using the commitment loss in our experiments as it led to model divergence.

To initialize the codebooks, we use a procedure similar to \texttt{kmeans++} initialization \cite{kmeansplusplus}. Additionally, we perform codebook reset \cite{codereset} using the same procedure to re-initialize infrequently used codes.

We divide memories into $g$ subspaces, and if needed, pad memories with zeros to ensure divisibility. Each subspace has $C$ codes. Therefore the storage requirement for each quantized vector is the number of subspaces multiplied by the number of bits required to represent each ID, which is the logarithm of the number of codes.
\vspace{-3pt}
\begin{equation*}
    S_{\lumenvq} = g \cdot \left\lceil \log_2{C} \right\rceil \cdot N
\end{equation*}

\section{Experiments}
\input{results/main_perf_vs_compression}
\input{results/ablations_table}

\subsection{Experimental setup}

\paragraph{Model configuration}
\lumenvq and \lumen are built on the T5.1.1 architecture \citep{t5} and implemented in JAX using Flax \citep{flax} and Flaxformer. All models fine-tune public T5.1.1 XXL checkpoints. We train FiD using the recipe from \citet{fid}.

The training setup for \lumen and \lumenvq is based on \citet{glimmer}. We initialize the memory encoder with the first 1 - $\liveprop$ proportion of layers from the T5 encoder and the live encoder with the last $\liveprop$ proportion of layers, where $\liveprop$ is the given proportion of live layers. We set $\liveprop = \frac{1}{3}$ in our main experiments.

We train and evaluate on a subset of knowledge-intensive task datasets from the KILT benchmark \citep{kilt}. We adopt the retrieval procedure from \citet{multikilt} and use GTR-Base model \citep{gtr} as the retriever. See Appendix~\ref{sec:appendix_experimental_setup} and \citet{glimmer} for details.

\subsection{Main results}
In our main experiments, we compress \lumen-XXL's 4096-dimensional memories using $g=256$ subspaces and $C=65536$ codes per subspace, allowing us to store code IDs in \texttt{int16} format. We need 512 bytes to store each token vector instead of 8192 bytes for the original memories. As a result, \lumenvq achieves a compression rate of 16 with minimal performance loss, as shown in Table~\ref{table:storage_size}.

\subsection{Quality-Compression rate trade-off}
We investigate the quality-compression tradeoff for \lumenvq by varying the number of subspaces.

We compare against several naive baselines; the first involves scaling down the model (e.g., LUMEN-XL or LUMEN-Large). This reduces $d$ from 4096 to 2048 or 1024, respectively. The second baseline, called LUMEN-Light, is inspired by the FiD-Light approach~\citep{fidlight}. In LUMEN-Light, we retain memories of the first K tokens, varying K from $\frac{1}{2}$ to $\frac14$ of the passage length, achieving compression rates of 2 and 4.

Figure~\ref{fig:main_perf_vs_compression} presents the performance results. Both baselines exhibit significant performance losses as compression rates increase. In contrast, the \lumenvq measure shows a gradual decline in performance, with a loss of approximately 0.2 performance points at a compression rate of 16.

\subsection{Ablations}
\label{sec:ablations}
We investigate if initializing VQ-VAE training from a fine-tuned \lumen model yields better results. The results in Table~\ref{table:ablations_table} show that fine-tuning \lumenvq from scratch achieves similar performance to initializing from a fine-tuned \lumen model.

We also analyze which model components benefit most from joint fine-tuning with VQ-VAE. Freezing the memory encoder during joint training, starting with a fine-tuned \lumen model, has little impact on performance. However, updating only VQ-VAE codes while freezing the entire model leads to decreased performance, indicating the model's need to adapt to decompression layer errors.

\section{Related work}
\label{section:related_work}

\paragraph{Memory models}
Retrieval augmentation can be computationally expensive due to the additional context that language models need to process. To mitigate this, memory models like LUMEN \citep{lumen}, GLIMMER \citep{glimmer}, and others \citep{readtwice,tome, memorizing, fidmemory, trime, qama, emat, unlimformer, lait} store pre-computed representations in memory. \memoryvq focuses on improving the storage requirements for memory-based models. While our experiments involve the \lumen \citep{lumen} model due to its strong performance, the method applies to a broader range of models.

\paragraph{Compression for late-interaction reranking}
\memoryvq focuses on compression for late-interaction memory models, while other works have explored compression for late-interaction reranking. For instance, SDR \citep{sdr} employs an autoencoder to reduce token representation dimensionality, followed by product quantization. BECR \cite{becr} utilizes locality-sensitive hashing for token representation compression. CQ \citep{cq} learns vector quantization parameters by treating codes as learnable weights and uses Gumbel-Softmax for differentiable nearest code determination. Finally, ColBERTv2 \citep{colbertv2} proposes a custom compression scheme combining PQ and integer quantization to handle reconstruction residuals.

\section{Conclusion}
\label{section:conclusion}
We introduced \modelname, a novel approach for reducing the storage requirements of memory-augmented language models without compromising performance. By employing VQ-VAE to compress token representations, we obtain a \lumen model with 16x compression, denoted as \lumenvq. Remarkably, \lumenvq maintains performance close to \lumen and \fid and benefits from \lumen inference speed-ups with sharply reduced storage cost. Using \memoryvq, memory augmentation is a practical solution for drastic inference speedups with extensive retrieval corpora.


\input{main.bbl}
\appendix

\section{Experimental setup}
\label{sec:appendix_experimental_setup}

\paragraph{Model configuration}
The original \lumen implementation employed a separate question encoder, but \citet{glimmer} showed we can re-use the memory encoder as long as it is fine-tuned.

\paragraph{Fine-tuning}
During fine-tuning, we utilize the Adafactor optimizer \citep{adafactor} with a constant learning rate of 0.0001, a batch size of 128, and a dropout rate of 0.1 for all tasks. When performing multi-task training, we uniformly sample from the tasks. We allocate 48 and 304 tokens for question and passage inputs, respectively. \lumenvq is using 0.999 as an EMA factor for code updates.

\paragraph{Data}
We train and evaluate on a subset of knowledge-intensive task datasets from the KILT benchmark \citep{kilt}. The datasets include question-answering datasets such as Natural Questions \citep{nq}, TriviaQA \citep{triviaqa}, and HotPotQA \citep{hotpotqa}, along with the fact verification dataset FEVER \citep{fever}, and the slot-filling datasets Zero Shot RE \citep{zeroshot} and T-REx \citep{trex}. To address imbalanced dataset issues, we apply the relevance filtering procedure introduced by \citet{multikilt}.

For the retrieval corpus, we use a Wikipedia dump provided by the KILT benchmark \url{http://dl.fbaipublicfiles.com/BLINK/enwiki-pages-articles.xml.bz2} containing approximately 4B tokens.

\paragraph{Retrieval}
We adopt the retrieval procedure introduced by \citet{multikilt}, where Wikipedia articles are segmented into chunks, each containing up to 200 words. The dense retriever, a pre-trained GTR-Base model \citep{gtr}, is utilized to identify the most relevant chunks for each query, with 20 retrieved passages for each query.

\section{Experiments}
\label{sec:appendix_experiments}

\subsection{Smaller codebook}
\input{results/perf_vs_compression_for_different_codebooks}
We study the effect of using a smaller codebook of size $C=4096$ instead of $C=65536$. Results in Figure~\ref{fig:perf_vs_compression_for_different_codebooks} show that using a smaller codebook has similar quality-compression trade-offs for lower compression rates but leads to worse trade-offs when we increase the compression rate.

\subsection{Can we compress code IDs even further?}
Integer data, like token IDs, might exhibit regularities, enabling additional data compression by using fewer bits for frequent patterns. For instance, applying standard compression tools like \texttt{gzip} or \texttt{zstd} to Wikipedia token IDs resulted in a compression factor of around 1.5. However, using the same tools on \lumenvq codes of Wikipedia passages yielded a more modest compression rate of 1.1.

Compression was performed independently on each subspace, with most subspaces being incompressible. Around 5\% of the subspaces showed compression rates ranging from 2 to 6. Notably, no compression was achieved when attempting to compress codes from all subspaces together.


\end{document}

%% file: custom_commands.tex
\newcommand{\modelname}{\textsc{memory-vq}\xspace}
\newcommand{\lumenvq}{\textsc{lumen-vq}\xspace}
\newcommand{\lumenvqshort}{\textsc{l-vq}\xspace}
\newcommand{\lumen}{\textsc{lumen}\xspace}

\newcommand{\memoryvq}{\textsc{memory-vq}\xspace}
\newcommand{\fid}{FiD\xspace}
\newcommand{\liveprop}{\alpha}

\definecolor{customblue}{rgb}{0, 0.46484375, 0.73046875} 
\definecolor{customred}{rgb}{0.9296875, 0.3984375, 0.46484375} 
\definecolor{customtangerine}{rgb}{0.95, 0.52, 0.0} 
\definecolor{customyellow}{rgb}{0.98, 0.85, 0.37} 
	
\definecolor{customgreen}{rgb}{0.18, 0.55, 0.34} 

\definecolor{flycolor}{rgb}{0, 0.46484375, 0.73046875} 
\definecolor{precolor}{rgb}{0.95, 0.52, 0.0} 

\definecolor{darkgrey}{rgb}{0.33203125,0.33203125,0.33203125} 

\pgfdeclareplotmark{redstar}{
    \node[star,star point ratio=2.25,minimum size=12pt,
          inner sep=0pt,draw=customred,solid,fill=customred] {};
}

%% file: results/main_table.tex
\begin{table}[ht!]
\centering
\begin{tabular}{lccc}
     & \textbf{FiD} & \textbf{LUMEN} & \textbf{L-VQ}\\
    \toprule
    &\multicolumn{3}{c}{\textit{Inference cost in TFLOPs}}\\
    Per sample & 28.0 & 12.5 & 12.5 \\
    \toprule
    &\multicolumn{3}{c}{\textit{Storage cost}}\\
    Per token & 2 bytes & 8 KB & 0.5 KB \\    
    For Wikipedia\protect  & 8 GB & 30 TB & 2 TB \\
    For 1T tokens& 2 TB & 7 PB & 0.5 PB \\
    \toprule
    &\multicolumn{3}{c}{\textit{KILT valid in \% exact match}}\\
    \textit{Average} & 72.80 & 72.66 & 72.42 \\
    \hline
    NaturalQuestions & 61.47 & 62.64 & 62.74 \\    
    TriviaQA & 83.40 & 82.84 & 82.61 \\
    FEVER & 93.47 & 92.77 & 92.18 \\    
    TREX & 83.58 & 83.78 & 83.42 \\    
    ZeroShot RE & 72.77 & 72.85 & 72.61 \\    
    HotpotQA & 42.09 & 41.09 & 41.00 \\
    \bottomrule
\end{tabular}
\caption{\textbf{Main results: \lumenvq (\lumenvqshort) nearly matches Fusion-in-Decoder in quality while benefiting from \lumen compute savings without impractical \lumen storage requirements.}}
\label{table:storage_size}
\end{table}





%% file: results/main_perf_vs_compression.tex
\begin{figure}[h!]
\centering
\begin{tikzpicture}[scale=1.0]
    \begin{axis}[
    scale only axis,
    width=0.8\columnwidth,
    height=0.7\columnwidth,            
    ylabel={KILT (exact match)},
    xlabel={Compression rate},
    mark=x,
    ymajorgrids=true,
    xmajorgrids=true,
    xminorticks=true,
    grid style=dashed,
    legend columns=1,
    legend cell align=left,
    legend pos={south east},
    ]
    \legend{\lumen, \lumenvq (main), \lumenvq, Scale down, \lumen-Light}

    \addplot[color=customblue,line width=3, dotted] table {
        1 72.661181
    	32 72.661181
    }; 
    
    \addplot[only marks, mark=redstar] coordinates {
        (16,72.357364)
    };
    
    \addplot[color=customred,mark=*,mark size=1pt,line width=2] table {
        4   72.66627
        8	72.580483
        16	72.357364
        20.5 72.157039
        23.5 72.00693
        32	71.827365
    };
    
    \addplot[color=customtangerine,line width=2] table {
    	2 71.692482
    	4 68.965358
    };    
    \addplot[color=customgreen,line width=2] table {
    	2 71.540976
    	4 69.559326
    }; 

    
\end{axis}
\end{tikzpicture}  
\caption{\textbf{\lumenvq achieves a strongly improved trade-off between performance and compression.} The plot shows average exact match on dev sets of KILT tasks as a function of compression rate. We compare \lumenvq with baselines \text{Scale down} (\lumen XL and \lumen Large) and \textit{\lumen-Light} (FiD-Light from \citet{fidlight} adapted for \lumen).}

\label{fig:main_perf_vs_compression}
\end{figure}
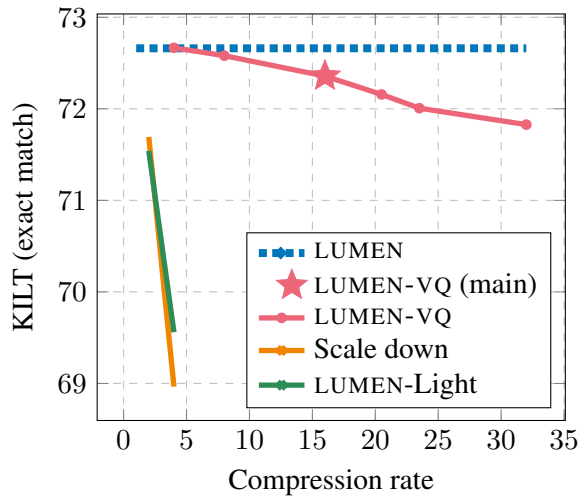

%% file: results/ablations_table.tex
\begin{table}[ht!]
\centering
\begin{tabular}{lccccccc}
     \textbf{Model} & \textbf{KILT, EM} \\
    \toprule
    \lumenvq & 72.43 \\
    \hline
    initialize from fine-tuned \lumen & 72.42 \\
    + freeze memory encoder & 72.33 \\
    + freeze whole model & 71.79 \\
    \bottomrule
\end{tabular}
\caption{Performance comparison of different approaches for initializing and training the \lumenvq.} 
\label{table:ablations_table}
\end{table}



%% file: results/perf_vs_compression_for_different_codebooks.tex
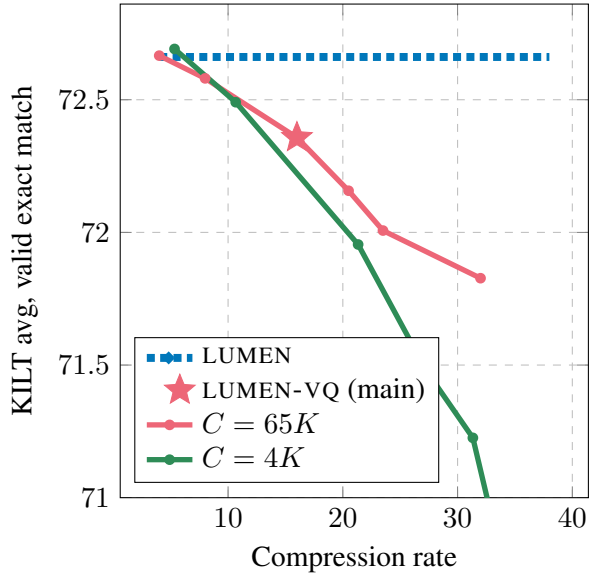
\begin{figure}[!ht]
\centering
\begin{tikzpicture}[scale=1.0]
    \begin{axis}[
    scale only axis,
    width=0.8\columnwidth,
    height=0.85\columnwidth,            
    ylabel={KILT avg, valid exact match},
    xlabel={Compression rate},
    mark=x,
    ymin=71,
    ymajorgrids=true,
    xmajorgrids=true,
    xminorticks=true,
    grid style=dashed,
    legend columns=1,
    legend cell align=left,
    legend pos={south west},
    ]
    \legend{\lumen, \lumenvq (main), $C=65K$, $C=4K$}

    \addplot[color=customblue,line width=3, dotted] table {
        4 72.661181
    	38 72.661181
    }; 
    \addplot[only marks, mark=redstar] coordinates {
        (16,72.357364)
    };
    
    \addplot[color=customred,mark=*,mark size=1pt,line width=2] table {
        4   72.66627
        8	72.580483
        16	72.357364
        20.5 72.157039
        23.5 72.00693
        32	71.827365
    };

    
    
    \addplot[color=customgreen,mark=*,mark size=1pt,line width=2] table {
        5.333333333 72.69146
        10.66666667 72.490699
        21.33333333 71.954442
        27.33333333 71.475149
        31.33333333 71.225569
        42.66666667 69.15676
    };        

\end{axis}
\end{tikzpicture}  
\caption{The plot shows average exact match on validation sets of KILT tasks as a function of compression rate. We compare \lumenvq with the codebook of size $C=65536$ and $C=4096$.}

\label{fig:perf_vs_compression_for_different_codebooks}
\end{figure}